# Modeling Car-Following Behavior on Urban Expressways in Shanghai: A Naturalistic Driving Study

Meixin Zhu[a,b], Xuesong Wang[a,c*], Andrew Tarko[d], Shou'en Fang[a,c]

[a]*Key Laboratory of Road and Traffic Engineering, Ministry of Education, Shanghai, 201804, China*
[b]*Department of Civil and Environmental Engineering, University of Washington, Seattle, WA 98195-2700, USA*
[c]*School of Transportation Engineering, Tongji University, Shanghai, 201804, China*
[d]*Lyles School of Civil Engineering, Purdue University, West Lafayette, Indiana, 47907, USA*
*Corresponding Author. Tel: +86-21-69583946. E-mail address: wangxs@tongji.edu.cn

**Abstract:**

Although car-following behavior is the core component of microscopic traffic simulation, intelligent transportation systems, and advanced driver assistance systems, the adequacy of the existing car-following models for Chinese drivers has not been investigated with real-world data yet. To address this gap, five representative car-following models were calibrated and evaluated for Shanghai drivers, using 2,100 urban-expressway car-following periods extracted from the 161,055 km of driving data collected in the Shanghai Naturalistic Driving Study (SH-NDS). The models were calibrated for each of the 42 subject drivers, and their capabilities of predicting the drivers' car-following behavior were evaluated.

The results show that the intelligent driver model (IDM) has good transferability to model traffic situations not presented in calibration, and it performs best among the evaluated models. Compared to the Wiedemann 99 model used by VISSIM®, the IDM is easier to calibrate and demonstrates a better and more stable performance. These advantages justify its suitability for microscopic traffic simulation tools in Shanghai and likely in other regions of China. Additionally, considerable behavioral differences among different drivers were found, which demonstrates a need for archetypes of a variety of drivers to build a traffic mix in simulation. By comparing calibrated and observed values of the IDM parameters, this study found that 1) interpretable calibrated model parameters are linked with corresponding observable parameters in real world, but they are not necessarily numerically equivalent; and 2) parameters that can be measured in reality also need to be calibrated if better trajectory reproducing capability are to be achieved.

## 1. Introduction

Over the last few decades, microscopic traffic simulation programs have become increasingly important tools for traffic and traffic safety engineering (e.g., capacity analysis, traffic impact studies, junction design, accident analysis, network analysis) (Ranjitkar et al., 2005). The cornerstones of microscopic traffic simulation are car-following models that describe the vehicle-by-vehicle following process in a traffic flow (Brackstone and McDonald, 1999). The performance of the car-following model used in simulation is one of the determinants of the traffic simulation validity.

Since the early investigation of car-following dynamics in 1953, numerous car-following models have been developed. The development and investigation of these models have been almost entirely based on experiments conducted in Western countries. However, drivers in different countries have different driving styles, drive different types of vehicles, and are subject to different traffic regulations as well as driving cultures (Daamen et al., 2013; Treiber and Kesting, 2013; Lindgren et al., 2008b). For example, Chinese drivers face the challenging driving environment of frequent lane changing, aggressive driving, considerable presence of large trucks, and omnipresent pedestrians, electric bikes, and bicycles (Huang et al., 2006; Lindgren and Chen, 2007). China also struggles with infrastructural issues such as inadequate road design and control, poor road maintenance, and insufficient road construction management (Lindgren et al., 2008a).

These influences may result in considerable differences in driving behavior and traffic operation (Daamen et al., 2013). Huang et al. (2006) found Chinese drivers to be more aggressive than drivers in the U.S.; for example, the Chinese tend to drive more offensively and to disobey traffic rules. Because car-following models are based on certain assumptions about driving behavior, a car-following model that performs well for drivers in Western countries may perform poorly when applied to drivers in developing countries.

With the rapid motorization in developing countries such as China, there is an increasing need to evaluate and to adjust, if needed, the existing car-following models applied to non-Western drivers. To address this need, Shanghai was chosen as a case study, and real-world driving data were collected in the Shanghai Naturalistic Driving Study (SH-NDS). The data collection procedure started in December 2012, and by December 2015 driving data had been collected for 60 drivers who drove 161,055 km in total. In the SH-NDS, driver behavior was observed as it occurred in the full context of real-world driving, and vehicle kinematic data (e.g., acceleration, velocity, position) were recorded continuously at high resolution. The availability of these detailed naturalistic driving data provides an unprecedented opportunity for investigating car-following behavior in China.

Car-following behavior on urban expressways is the focus of the current study. Urban expressways form the backbone of city road networks, carrying a disproportionately large percentage of city traffic. In Beijing, for example, major urban expressways account for only 8% of the total road network length, but carry nearly 50% of the traffic flow; in Shanghai, urban expressways are only 5% of the city's road network, but they bear more than 35% of the city's traffic (Chen et al., 2014). Although car-following behavior on non-freeway arterials should also be studied, the presence of intersections introduces additional variables into car-following behavior that are beyond the scope of this study. Therefore, urban expressway data from the SH-NDS were used in the current study to calibrate and cross-compare five different car-following models, with the final aims of attaining better insight into the car-following behavior of Chinese drivers and determining the best model for use in Shanghai.

## 2. Literature review

### 2.1. Car-following models

A car-following model describes the movements of a following vehicle (FV) in response to the actions of the lead vehicle (LV). The first car-following models (Pipes, 1953; Chandler et al., 1958) were proposed in the middle 1950s. Since then, a large number of models were developed, for example, the Gazis-Herman-Rothery model (Gazis et al., 1961), the intelligent driver model (Treiber et al., 2000), the optimal velocity model (Bando and Hasebe, 1995), and the models proposed by Helly (1959), Gipps (1981), and Wiedemann (1974). For a review and historical development of the subject, one should consult Brackstone and McDonald (1999), Olstam and Tapani (2004), Panwai and Dia (2005), and Saifuzzaman and Zheng (2014). Generally, car-following models can be divided into five categories: stimulus-based, safety-distance, desired measures, optimal velocity, and psycho-physical models.

The main idea of a stimulus-based model is that the acceleration of a following vehicle is determined by the driver's reaction to the speed and position differences of the vehicle in front (May, 1990). The General Motors models are some of the best known stimulus-based models. They have been developed since the late 1950s, with one of their latest modifications proposed by Ozaki (Ozaki, 1993).

The safety-distance models are based on the idea that the driver of a following vehicle would adopt a speed and maintain a distance such that he/she can bring the vehicle to a safe stop should the vehicle in front brake to a sudden stop. The Gipps model (Gipps, 1981) is based on the safety-distance idea.

The desired measures models assume that a driver has a preferred situation represented by certain measures (e.g., following distance and following speed) and a driver continuously attempts to eliminate the difference between the actual situation and the preferred. The intelligent driver model (IDM) (Treiber et al., 2000) is one of the most widely used models using desired measures.

The optimal velocity models assume that each following vehicle has an optimal safe velocity that depends on the distance from the lead vehicle. The acceleration of the following vehicle can be determined according to the difference between the actual velocity and the optimal velocity (Saifuzzaman and Zheng, 2014). Closely related to the optimal velocity model is the full velocity difference (FVD) model (Jiang et al., 2001).

The psycho-physical models suggest that a driver's behavior would vary depending on the traffic state he/she is in, such as whether the driver is in the free-flow condition, approaching the vehicle in front, following the vehicle in front, or braking. The boundary conditions defining the different states are usually expressed as a combination of relative speed and relative distance to the lead vehicle (Wiedemann, 1974).

### 2.2. Calibrating car-following models with microscopic data

Calibration involves finding a set of model parameters that minimize the difference between the values of the simulated and observed variables (Kesting and Treiber, 2008).

Previous studies have calibrated car-following models with microscopic data collected from observations on existing roads with normal traffic or on field experimental tracks with the participation of selected drivers. Field experimental track studies may suffer bias due to the lack of real-world driving conditions (Ranjitkar et al., 2005; Kesting and Treiber, 2008), but studies using observations on existing roads depend on driving data from anonymous drivers and thus lack driver-specific information (NGSIM, 2005; Montanino and Punzo, 2015; Chen et al., 2010). Recent technological advances in microscopic data-collection methods, however, have permitted the unobtrusive collection of driver-specific information known as naturalistic driving data. Naturalistic data represent drivers' natural behavior, including complete trajectories and individual driver characterization. These data are found to be superior over the aforementioned data types as they provide unique insight into the car-following behavior of diverse drivers (Sangster et al., 2013). Thus, the use of naturalistic driving data is continually increasing (Abbas et al., 2011; Higgs and Abbas, 2014; Higgs, 2011; Sangster et al., 2013).

Once the data collection technique is determined, the method of calibration is considered. Investigation of methodological studies on calibrating car-following models (Daamen et al., 2013; Treiber and Kesting, 2013; Hollander and Liu, 2008; Ossen and Hoogendoorn, 2008; Ciuffo and Punzo, 2010) identifies three calibration components: (1) measure of performance (MoP), (2) goodness of fit (GoF), and (3) optimization algorithm.

The MoP is the variable that represents the car-following behavior. Its values observed in the field are compared to the values produced by the model being calibrated. Although the space headway (distance from the FV's front bumper to the LV's front bumper) and speed of the following vehicle are two commonly used MoPs, Punzo and Montanino (2016) demonstrated that spacing (FV's front bumper to LV's rear bumper, also referred to in this study as *gap* and *following distance*) allows more efficient and robust calibration.

GoF measures the difference between the observed and simulated MoPs. The root mean square percentage error (RMSPE), which represents the relative errors, is a widely used GoF:

$$RMSE(x,y) = \sqrt{\frac{1}{N}\sum_{i=1}^{N}(x_i - y_i)^2} \qquad (1)$$

where $x$ and $y$ are the simulated and observed MoPs, respectively, while $N$ is the total number of observations and $i$ is an observation index. For a detailed discussion of different GoFs, refer to Ciuffo et al. (2012).

Three optimization algorithms are widely used to calibrate car-following models: (1) downhill simplex, (2) genetic algorithm (GA), and (3) OptQuest/Multistart. The downhill simplex is suitable for unconstrained nonlinear optimization, while the GA and OptQuest/Multistart can solve both constrained and unconstrained global optimization problems. Among the three algorithms, GA is the most widely used because it can avoid local minima and reach the global optimum with a stochastic (random) global search method (Saifuzzaman et al., 2015).

## 3. Car-following models investigated in this study

The current study compared five car-following models: Gazis-Herman-Rothery, Gipps,

intelligent driver, full velocity difference, and Wiedemann. The chosen models are representative of the model types reviewed in Section 2.1. A summary of the parameters for estimation for the first four evaluated models can be found in Table 2 below in Section 6. Due to a large number of parameters in the Wiedemann model, these parameters are presented in Appendix A in Table A.1.

*3.1. Gazis-Herman-Rothery (GHR) model*

The GHR model, also known as the General Motors (GM) model, is the best known stimulus-based model and is the result of research spanning from the late 1950s until the middle 1960s (Chandler et al., 1958; Gazis et al., 1961). A discussion on the development of the GM models along with typical parameter values can be found in May (1990), among others. The fifth and latest formulation of the model, sometimes referred to as the GM-5, is the generalization of the previous four forms:

$$a_n(t) = \alpha V_n(t)^\beta \frac{\Delta V_n(t-\tau_n)}{\Delta X_n(t-\tau_n)^\gamma} \tag{2}$$

where $a_n(t)$ is the acceleration of the following vehicle at time $t$, $\Delta V_n(t-\tau_n)$ is the speed difference between the following vehicle and the preceding vehicle at time $(t-\tau_n)$, $\Delta X_n$ is the spacing from the preceding vehicle, $\tau_n$ denotes the reaction time, and $\alpha$, $\beta$, and $\gamma$ are parameters.

*3.2. Gipps model*

The Gipps model (1981) is a typical safety-distance model. The model includes two modes of driving: free-flow and car-following. The driver chooses the lower of the speeds obtained from the free-flow and car-following modes, as shown in Equation (*3*):

$$V_n(t+\tau_n) = \min \begin{cases} V_n(t) + 2.5\tilde{a}_n\tau_n\left(1-V_n(t)/\tilde{V}_n\right)\left(0.025 + V_n(t)/\tilde{V}_n\right)^{1/2} \\ -\tilde{b}_n\tau_n + \sqrt{\tilde{b}_n^2\tau_n^2 + \tilde{b}_n\left[2(\Delta X_n(t) - S_{n-1}) - V_n(t)\tau_n + \frac{V_{n-1}(t)^2}{\hat{b}}\right]} \end{cases} \tag{3}$$

where $\tilde{a}_n$ is the desired acceleration, $\tilde{b}_n$ is the desired deceleration, $S_{n-1}$ is the effective length of the lead vehicle (length of the vehicle plus a safety distance into which the following vehicle is not willing to intrude even when at rest), $V_n$ and $V_{n-1}$ are the speeds of the following and lead vehicle respectively, $\hat{b}$ is an estimate of the deceleration applied by the lead vehicle, and $\tilde{V}_n$ is the desired speed of the following vehicle. The Gipps model has been used in many simulation tools, including the commonly used AIMSUN ®.

## 3.3. Intelligent driver model (IDM)

The IDM (Treiberet al., 2000) is a typical desired measures model. It considers both the desired speed and the desired following distance, as defined in Equation (*4*):

$$a_n(t) = a_{max}^{(n)}\left[1-\left(\frac{V_n(t)}{\tilde{V}_n(t)}\right)^{\beta} - \left(\frac{\tilde{S}_n(t)}{S_n(t)}\right)^2\right] \tag{4}$$

where $a_{max}^{(n)}$ is the maximum acceleration/deceleration of the following vehicle $n$, $\tilde{V}_n$ is the desired speed, $S_n$ is the spacing between the two vehicles measured from the front edge of the following vehicle to the rear end of the lead vehicle ($S_n = \Delta X_n - L_n$; where $L_n$ is vehicle length), $\tilde{S}_n$ is the desired spacing, and $\beta$ is a parameter. The desired following distance in IDM is dependent on several factors: speed, speed difference ($\Delta V$), maximum acceleration ($a_{max}^{(n)}$), comfortable deceleration ($a_{comf}^{(n)}$), minimum spacing at standstill ($S_{jam}^{(n)}$), and desired time headway ($\tilde{T}_n$). Mathematically, the desired following distance can be calculated using Equation (*5*):

$$\tilde{S}_n(t) = S_{jam}^{(n)} + \max\left(0, V_n(t)\tilde{T}_n(t) - \frac{V_n(t)\Delta V_n(t)}{2\sqrt{a_{max}^{(n)} a_{comf}^{(n)}}}\right) \tag{5}$$

## 3.4. Full velocity difference (FVD) model

The FVD model (Jiang et al., 2001) is closely related to the optimal velocity model developed by Bando et al. (1995). The acceleration function consists of a term proportional to a gap-dependent optimal velocity $V_n^*(\Delta X_n(t))$ and a term that takes into account velocity difference $\Delta V_n$ as a linear stimulus:

$$a_n(t) = \alpha\left[V_n^*(\Delta X_n(t)) - V_n(t)\right] + \lambda(\Delta V_n(t)) \tag{6}$$

$$\lambda = \begin{cases} \lambda_0: & \Delta X_n(t) \leq s_c \\ 0: & \Delta X_n(t) > s_c \end{cases} \tag{7}$$

where $\alpha$ and $\lambda_0$ are constant sensitivity coefficients, $s_c$ is a threshold between following and free driving, and $V_n^*(\Delta X_n(t))$ is the optimal velocity depending on the headway $\Delta X_n$ to the lead vehicle, which can be defined as:

$$V_n^*(\Delta X_n(t)) = \frac{V_0}{2}\left[\tanh\left(\frac{\Delta X_n(t) - L_{n-1}}{b} - \beta\right) - \tanh(-\beta)\right] \tag{8}$$

where $V_0$ is the desired speed of the following vehicle, and $b$ and $\beta$ are parameters.

*3.5. Wiedemann model*

The Wiedemann 74 (Wiedemann, 1974) and Wiedemann 99 (VISSIM, 2012) are typical psycho-physical models. They are building blocks of the widely used microscopic traffic simulation tool, VISSIM®. In these models, the psychological and physical aspects of driving behavior form the basis for distinguishing four discrete driving regimes: (1) free-flow, (2) approaching slower vehicles, (3) car-following near steady-state equilibrium, and (4) critical situations requiring braking action. In each of these regimes, different acceleration functions $a(s, v, \Delta v)$ apply. The boundaries between the regimes are given by nonlinear equations of the form $f(s, v, \Delta v) = 0$ defining curved areas in three-dimensional state space $(s, v, \Delta v)$ spanned by the exogenous variables (Fellendorf and Vortisch, 2010). The Wiedeman 99 model was investigated in this study because it is more suitable than the 74 model version for modeling expressway car-following behavior. Detailed information on acceleration functions and boundary equations of the Wiedemann 99 model can be found in Table A.1 in Appendix A1.

## 4. Data collection and preparation

*4.1. Shanghai Naturalistic Driving Study*

The data used in the current study were collected in the Shanghai Naturalistic Driving Study (SH-NDS) jointly conducted by Tongji University, General Motors (GM), and the Virginia Tech Transportation Institute (VTTI). The SH-NDS aimed to learn more about the vehicle use, vehicle handling, and safety awareness of Chinese drivers.

Five GM light vehicles equipped with Strategic Highway Research Program 2 (SHRP2) NextGen data acquisition systems (DAS) were used to collect real-world driving data. The three-year data collection procedure started in December 2012 and ended in December 2015. Driving data were collected daily from 60 licensed Shanghai drivers who, all together, traveled 161,055 km during the study period. The 60 participants were randomly sampled from the population of licensed Shanghai drivers; the distribution of gender, age, and driving experience of the sample accords with that of the general Chinese driving population.

The DAS uses an interface box to collect vehicle controller area network (CAN) data, an accelerometer for longitudinal and lateral acceleration, a radar system that measures range and range rate to the lead vehicle and the vehicles in adjacent lanes, a light meter, a temperature/humidity sensor, a global positioning system (GPS) sensor, and four synchronized camera views to validate the sensor-based findings (Fitch and Hanowski, 2012).

As shown in Fig. 1, the four camera views monitor the driver's face, the forward roadway, the roadway behind the vehicle, and the driver's hand movements. The data collection frequency ranges from 10 to 50 Hz. The DAS automatically starts when the vehicle's ignition is turned on, and automatically powers down when the ignition is turned off.

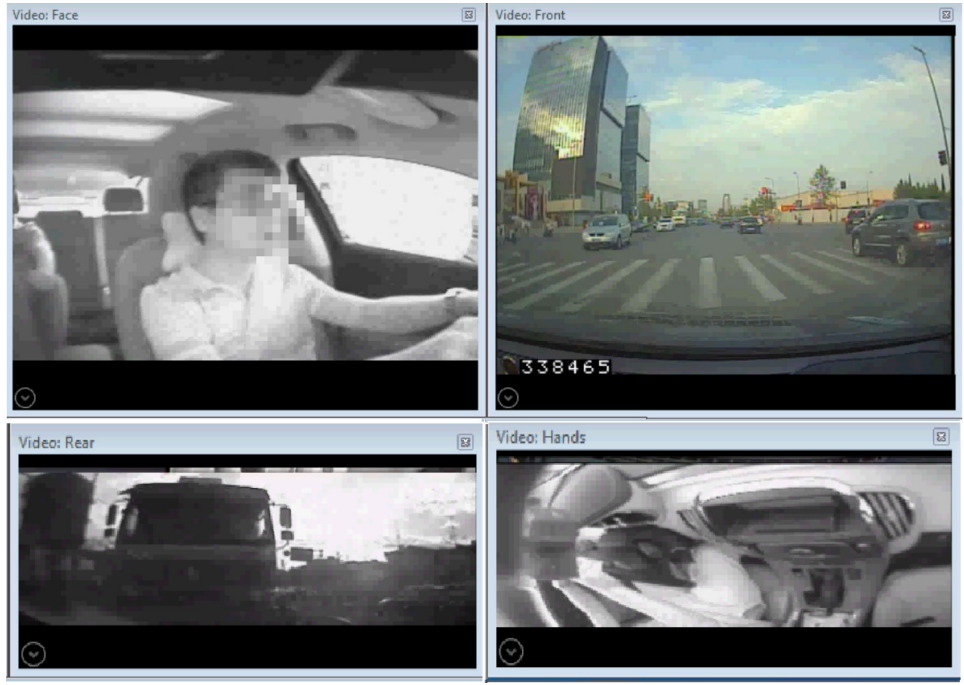

Fig. 1. Four camera views from the SH-NDS.

*4.2. Car-following periods extraction*

Car-following periods were automatically extracted for this analysis from the massive volume of collected data by applying a car-following filter, an iterative process where initial criteria and thresholds follow similar work by Ervin et al. (2005) and Higgs and Abbas (2013). After the potential car-following periods were flagged by the filter, the corresponding video material was reviewed to adjust the criteria and thresholds accordingly. A detailed description of determining car-following extraction criteria is presented in Appendix B.

As shown in Fig. 2, a car-following period was ultimately extracted if the following criteria were met simultaneously:
- Radar target's identification number >0 and remained constant: this criterion guaranteed that the same LV was being detected;
- Range <120 m: this criterion eliminated free-flow traffic conditions;
- Lateral distance <2.5 m: this criterion guaranteed that the following and lead vehicles were driving in the same lane; and
- Duration of car-following period >15 s: this criterion guaranteed that the car-following persisted long enough to be analyzed.

The results of the automatic extraction process were confirmed by an analyst viewing the extracted video material.

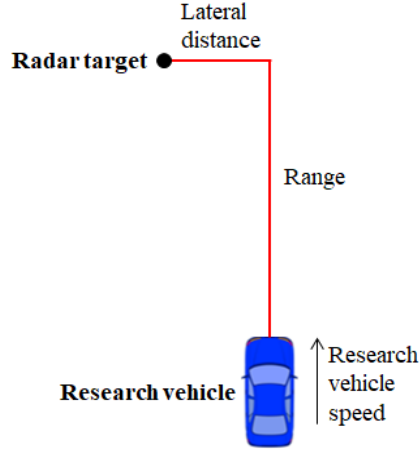

Fig. 2. Radar target's position and motion with respect to the research vehicle.

*4.3. Car-following periods analyzed*

To limit the focus of this study to car-following behavior on urban expressways, the roadway type of a car-following period was identified by viewing the corresponding video clip. As 42 of the original 60 drivers accounted for 97% of the urban expressway car-following periods, only these 42 drivers were investigated. Their ages ranged from 25 to 60 with a mean value of 38.83; and their driving experiences ranged from 1 to 23 years, with a mean value of 8.17 years. Among the 42 drivers, 20% were female. This female ratio is in line with the male-female disparity in China—female drivers accounted for 23.48% of all drivers in 2014 (Ministry of Public Security, China, 2015). For each of the 42 drivers, 50 car-following periods on urban expressways were randomly selected for investigation. The total 2,100 selected periods represented 863 minutes of driving.

## 5. Calibration and validation methodology

*5.1. Objective function*

The current study based its calibration on the more robust and efficient inter-vehicle spacing (FV's front bumper to LV's rear bumper) recommended by Punzo and Montanino (2016). To enable comparison of the results with a similar study, a variant of the root mean square percentage errors (RMSPE) of spacing used in Sangster et al. (2013) was adopted as the objective function:

$$\text{RMSPE of spacing} = \sqrt{\frac{\sum_{i=1}^{N}(S_i^{sim} - S_i^{obs})^2}{\sum_{i=1}^{N}(S_i^{obs})^2}} \qquad (9)$$

where $i$ denotes the observations with a sampling frequency of 10 Hz, $S_i^{sim}$ is the $i$th modeled spacing, $S_i^{obs}$ is the $i$th observed spacing, and $N$ is the number of observations.

## 5.2. Parameter constraints, and collision penalty

Table 2 (Section 6) and Table A.1 (Appendix) give an overview of the behavioral parameters of the investigated models and the bounds of those parameters. It should be noted that, in the simulation, some combinations of parameter values may lead to apparent instances of collision. This situation indicates poor performance of the models because it is contradictory to the reality that no collision was observed in any of the analyzed car-following periods. To make these combinations unattractive solutions to the optimization algorithm, a large crash penalty was included in the objective function, which is the standard procedure of numerical optimization (Kesting and Treiber, 2008).

## 5.3. Integration scheme

The forward Euler method was used to solve for the vehicle position and speed, as shown in Equations (10) and (11):

$$V_n(t+\Delta T) = V_n(t) + a_n(t) \cdot \Delta T \tag{10}$$

$$X_n(t+\Delta T) = X_n(t) + V_n(t) \cdot \Delta T \tag{11}$$

where $V$, $a$, $X$ are speed, acceleration, and position of the vehicle respectively, and $\Delta T$ is the update time step, set as 0.1s in the current study. Backward movements were removed by setting any negative speed to zero.

## 5.4. Genetic algorithm

GA was implemented to find the optimum values of the model parameters, which proceeded as follows:

1) A population consisting of N individuals was initialized, with each individual representing one parameter set of the car-following model;
2) The fitness of each individual in the population was determined via a predefined objective function;
3) Crossovers between randomly selected individual pairs (parents) and mutations within randomly selected individuals were implemented to produce individuals of the next generation (children); and
4) Steps 2 and 3 were repeated until the termination criteria were satisfied.

The Genetic Algorithm Toolbox in MATLAB® was used in this study, and the GA settings are described in Table 1. The relevant GA parameters for calibration of the GHR, Gipps, IDM, and FVD models were specified as follows: population size 300, maximum number of generations 300, and number of stall generations 100. As the number of parameters in the Wiedemann model (13 parameters) was significantly larger than in other models, the population size, maximum number of generations, and number of stall generations were set as 500, 1300, and 150, respectively.

The algorithm calculated the average relative weighted change in the fitness function value over the stall generations; if the change was less than function tolerance ($1 \times 10^{-6}$ in this study), the algorithm stopped. The maximum number of generations controlled the number of iterations.

As GA uses a stochastic process, it finds slightly different solutions in each optimization run. To find a solution closer to the global optimum, the optimization process was repeated 12 times for each driver, and the set of parameters with the minimum error (i.e., RMSPE) was selected.

Table 1 Description of the genetic algorithm settings (MathWorks, 2016).

| Algorithm setting | Method used | Method description |
| --- | --- | --- |
| Fitness scaling | Rank | Scales the raw scores based on the rank of each individual instead of its score. The rank of an individual is its position in the sorted scores. |
| Parents* selection | Stochastic uniform | Lays out a line in which each parent corresponds to a section of the line in a length proportional to its scaled value. The algorithm moves along the line in steps of equal size. At each step, the algorithm allocates a parent from the section on which it lands. |
| Children reproduction | Elite and crossover | A certain number of children are the elites of the parents (individuals who have the top 5% fitness values). The rest of the children are produced by crossover and mutation operations. |
| Mutation | Gaussian | Adds a random number taken from a Gaussian distribution with mean 0 to each entry of the parent vector. |
| Crossover | Scatter | Creates a random binary vector and selects the genes where the vector is a 1 from the first parent and the genes where the vector is a 0 from the second parent, and combines the genes to form the child. |

*To create the next generation, the genetic algorithm selects certain individuals in the current population, called **parents**, and uses them to create individuals in the next generation, called **children**.

### 5.5. Calibration with synthetic data

Before calibrating the models, the performance of the proposed calibration process was tested with the so-called "synthetic data" generated with each investigated model, as proposed by Punzo et al. (2012). For example, the parameters used in the GHR model were set as: $\tau_n = 1$, $\alpha = 1$, $\beta = 1$, and $\gamma = 1$, and the synthetic car-following data were generated. Then, the calibration process was implemented, resulting in optimal model parameters of $\tau_n = 1.00$, $\alpha = 1.19$, $\beta = 0.90$, and $\gamma = 0.99$. The calibration error (i.e., RMSPE) was very small (0.003); hence, it could be assumed that the proposed calibration process would be capable of finding the optimum model parameters if applied to actual data.

### 5.6. Calibration and validation workflow

The car-following models were calibrated at an individual-driver level, i.e., the calibration process was repeated for each driver, each of whom had his/her own set of parameters. The focus on calibrating individual drivers supports an idea of simulating a traffic flows as a mix of drivers with their own specific car-following behaviors rather than using average drivers that represent groups of drivers.

Cross-validation can minimize the impact of random factors and make full use of the data. To avoid biases from the selected calibration dataset and to perform a robust evaluation of the models, a five-fold cross-validation was applied to each driver. As shown in Fig. 3, the 50 car-following periods for each driver were randomly divided into five equal subsets; therefore, the model calibration and validation processes were repeated five times. In each iteration, four subsets were used to calibrate the car-following model and the remaining subset was used to conduct intra-driver validation. Upon completion, the calibration and validation errors from the five iterations were averaged to get the final performance measures of the model.

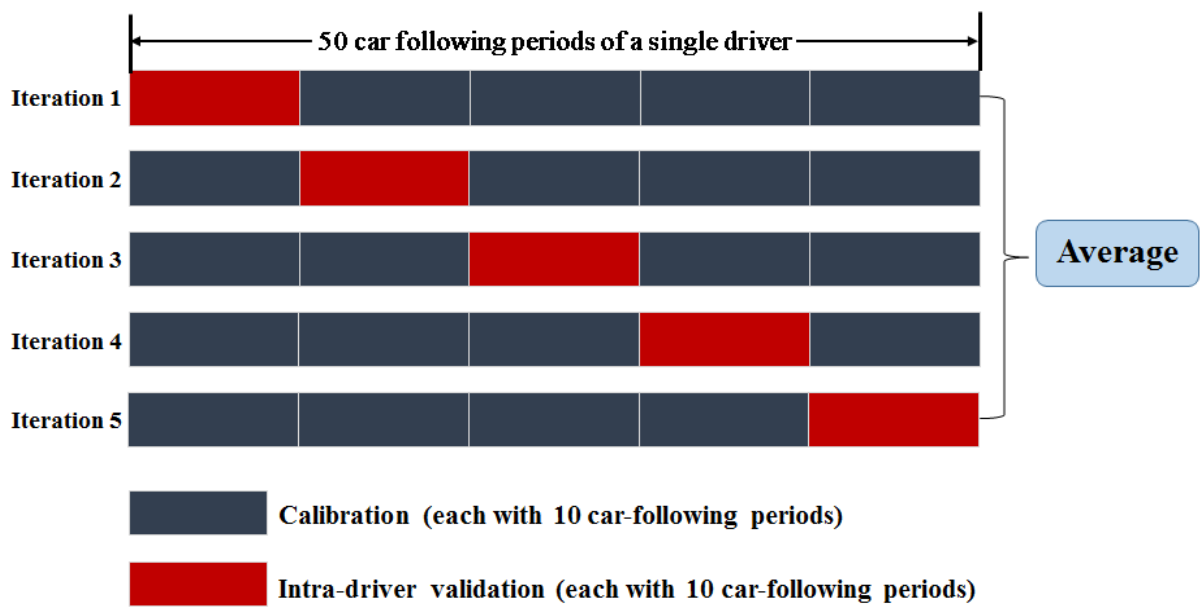

Fig. 3. Schematic representation of five-fold cross-validation iterations for a single driver involved in 50 car-following periods.

The calibration and validation proceeded as follows for each driver:
1) Forty (80%) of the 50 car-following periods were randomly selected as the calibration dataset based on the fold-selection procedure described above.
2) Using the calibration dataset, the car-following models were calibrated with GA based on the spacing between the lead and following vehicles, the outputting model parameters, and the spacing calibration errors.
3) The remaining 10 car-following periods of the driver being calibrated were used for validation, and the RMSPEs of both the FV speed and spacing were recorded to determine how the model would perform while considering differing measures of performance.

## 6. RESULTS AND ANALYSIS

*6.1. Estimates of model parameters*

The five car-following models were independently calibrated and validated five times for each of the 42 drivers. Thus, 5×42 sets of optimal parameters, calibration errors, and validation errors (RMSPE) were obtained for each model. The mean, standard deviation, and range (5th and 95th percentiles) of the estimated parameters for the investigated models are shown in Table 2 below (and Appendix Table A.1 for Wiedemann parameters). The intra-driver standard deviation of model parameters measures parameter variability between different car-following periods for the same driver, and is obtained by investigating parameters from different iterations of cross-validation; while the inter-driver standard deviation measures parameter variability between different drivers. These parameters provide guidance on traffic simulation, the cumulative distribution curves of the parameter estimates are presented in Appendix C (Fig. C.1 through Fig. C.5).

It may be observed in the estimated parameters for the investigated models in Table 2, that some of these parameters with the same meaning (e.g., reaction time) have different values across the various models. This might be due to parameter values being just the outcome of an optimization: they are the combination of values that minimizes the objective function, and the value of a certain parameter is affected by other parameters within the same model. Because different models have different parameter sets, it is not surprising that parameters with the same meaning should differ in value among the models.

Table 2 Summary of the model parameters and their estimates (aggregated across the five iterations of cross-validation and the 42 drivers).

| Model | Parameter (unit) | Short description | Bounds | Mean | Median | Std. dev. | 5% | 95% |
|---|---|---|---|---|---|---|---|---|
| GHR | $\alpha^a$ | Constant sensitivity coefficient | [0, 60] | 12.2124 | 8.3527 | 10.0015 | 1.8209 | 29.4708 |
| | $\beta^b$ | Sensitivity to FV speed | [-10, 10] | 0.5987 | 0.5891 | 0.4545 | -0.0532 | 1.3111 |
| | $\gamma^b$ | Sensitivity to space headway | [0, 10] | 1.5294 | 1.5047 | 0.5658 | 0.7171 | 2.5271 |
| | $\tau_n$ (s)$^c$ | Reaction time | [0.3, 3] | 0.6476 | 0.5000 | 0.4346 | 0.3000 | 1.4000 |
| Gipps | $\tilde{a}_n$ (m/s$^2$)$^c$ | Maximum desired acceleration of FV | [0.1, 5] | 1.4355 | 0.8563 | 1.4300 | 0.3749 | 4.9691 |
| | $\tilde{b}_n$ (m/s$^2$)$^c$ | Maximum desired deceleration of FV | [0.1, 5] | 1.2146 | 1.1379 | 0.7131 | 0.1862 | 2.3736 |
| | $S_{n-1}$ (m)$^c$ | Effective length of LV | [5, 15] | 5.6204 | 5.4207 | 0.5665 | 5.1042 | 6.6373 |
| | $\hat{b}$ (m/s$^2$)$^c$ | Maximum desired deceleration of LV | [0.1, 5] | 1.1145 | 1.0361 | 0.6441 | 0.1791 | 2.2829 |
| | $\tilde{V}_n$ (km/h)$^c$ | Desired speed of FV | [1, 150] | 90.6106 | 83.2725 | 22.0587 | 62.1729 | 147.682 |
| | $\tau_n$ (s)$^c$ | Reaction time | [0.3, 3] | 1.2214 | 1.2000 | 0.4831 | 0.5000 | 2.0400 |
| IDM | $a_{max}^{(n)}$ (m/s$^2$)$^c$ | Maximum acceleration/deceleration of FV | [0.1, 5] | 0.8418 | 0.8088 | 0.2775 | 0.5024 | 1.3222 |
| | $\tilde{V}_n$ (km/h)$^c$ | Desired speed of FV | [1, 150] | 105.6281 | 101.9284 | 24.7375 | 74.0000 | 150.000 |
| | $\beta^a$ | Acceleration exponent | [1, 40] | 3.5238 | 1.5000 | 3.6240 | 1.0000 | 10.0000 |
| | $a_{comf}^{(n)}$ (m/s$^2$)$^c$ | Comfortable deceleration of FV | [0.1, 5] | 0.8150 | 0.6123 | 0.7251 | 0.1177 | 2.3618 |
| | $S_{jam}^{(n)}$ (m)$^c$ | Gap at standstill | [0.1, 10] | 1.5554 | 1.3812 | 0.9633 | 0.1000 | 3.4247 |
| | $\tilde{T}_n$ (s)$^d$ | Desired time headway of FV | [0.1, 5] | 0.8780 | 0.9459 | 0.3343 | 0.3555 | 1.4319 |
| FVD | $\alpha^d$ | Constant sensitivity coefficient | [0.05, 20] | 0.0626 | 0.0500 | 0.0334 | 0.0500 | 0.1203 |
| | $\lambda_0^d$ | Sensitivity to relative speed | [0, 3] | 0.7018 | 0.6402 | 0.2635 | 0.4287 | 1.1883 |
| | $V_0$ (km/h)$^d$ | Desired speed of FV | [1, 252] | 120.2915 | 100.7714 | 45.7989 | 71.3982 | 231.161 |
| | $b^d$ | Interaction length | [0.1, 100] | 19.3901 | 16.6407 | 12.1905 | 4.8034 | 44.8287 |
| | $\beta^d$ | Form factor | [0.1, 10] | 1.0776 | 0.7802 | 1.5830 | 0.1500 | 3.5889 |
| | $S_c$ (m) | Max following distance | [10 120] | 46.9134 | 42.3362 | 25.1691 | 20.0623 | 106.183 |

Source: $^a$Sangster et al. (2013), $^b$Brackstone and McDonald (1999), $^c$Saifuzzaman et al. (2015), $^d$Kesting and Treiber (2008).

## 6.2. Calibration and validation errors

The empirical cumulative distribution functions (c.d.f.) of the calibration and intra-driver validation errors were calculated with Equation (*9*) and are shown in Fig. 4. The closer the c.d.f. was to the top left corner, the better the model's performance. It is obvious in Fig. 4 that the GHR model performed the worst in the calibration phase, and the FVD and IDM performed best. In the validation phase, the IDM performed the best no matter which type of performance measure (FV speed or spacing) was applied.

$$F_\varepsilon(e) = \frac{\text{Number of drivers with error } \varepsilon < e}{\text{Total number of drivers}} \qquad (9)$$

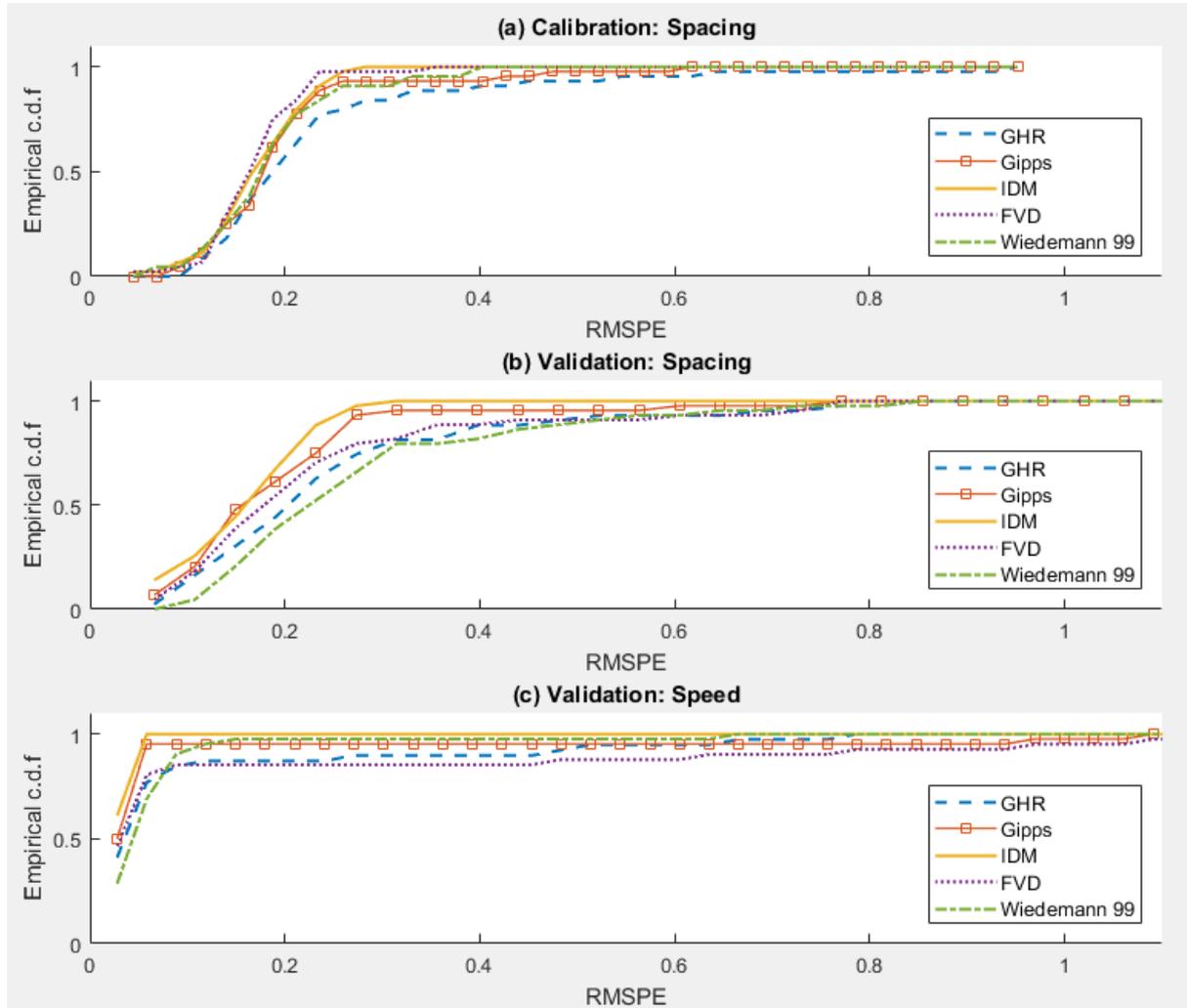

Fig. 4. Empirical c.d.f. of (a) calibration errors, (b) spacing validation errors, and (c) FV speed validation errors for the five car-following models.

Fig. 5 presents the mean values and respective standard deviations of the calibration and validation errors for the five models. Generally speaking, the RMSPE of FV speed is lower than that of spacing in the same validation phase. The IDM performed the best with validation data, and almost as well as the FVD model with calibration data. The IDM's robust performance is demonstrated by its smallest standard deviation of error in both the calibration and validation phases.

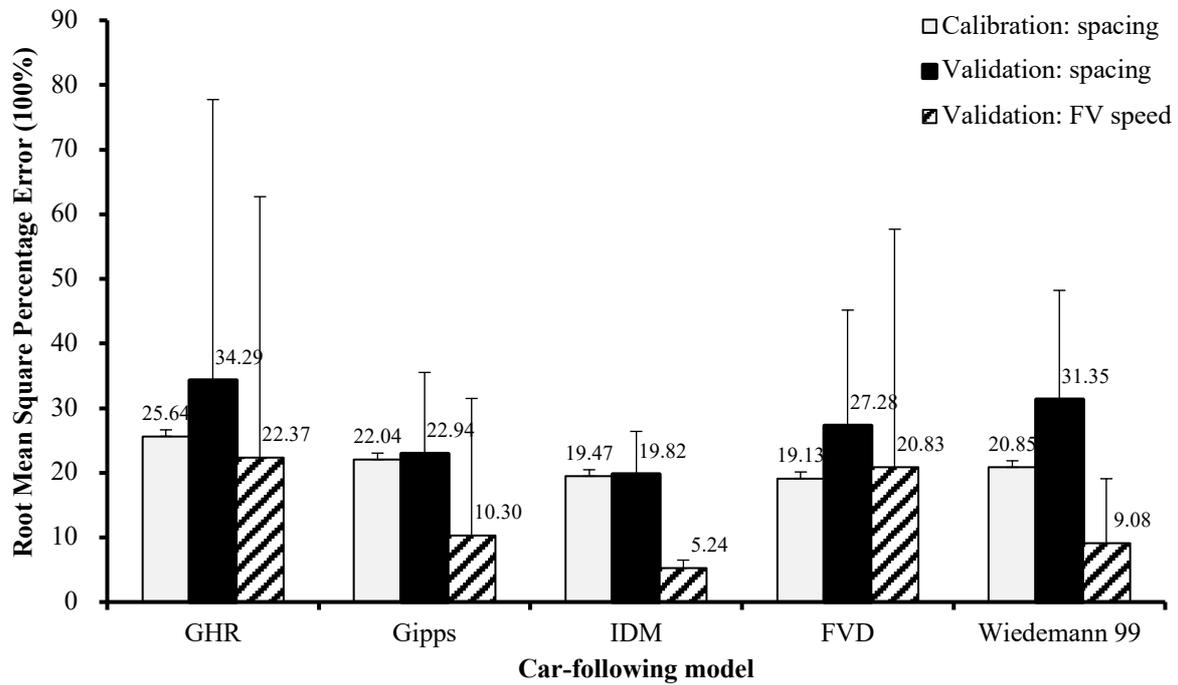

Fig. 5. Mean values and respective standard deviations of calibration and validation errors for the five models (averaged across the 42 drivers and the five iterations of cross-validation).

The numbers of collisions that occurred are shown in Table 3. The Gipps and IDM demonstrated good performance with no occurrences of collision, while the GHR, FVD, and Wiedemann 99 models all produced collisions, especially in the validation phase.

Table 3 Numbers of collisions and backward movements in calibration and validation phases.

| Model | Collision | |
|---|---|---|
| | Calibration | Validation |
| GHR | 6 | 40 |
| Gipps | 0 | 0 |
| IDM | 0 | 0 |
| FVD | 1 | 50 |
| Wiedemann 99 | 1 | 4 |

To summarize, the IDM had the overall best performance, considering that it had the following: 1) the lowest mean error in the calibration and validation phases, 2) the smallest standard deviation of error in both the calibration and validation phases, and 3) no occurrences of collision in the calibration and validation phases.

## 7. Summary and discussion

The study presented in this paper aims to identify car-following models suitable for use in Shanghai. Using 2,100 urban expressway car-following periods extracted from the Shanghai Naturalistic Driving Study (SH-NDS) database, five different car-following models were calibrated and validated and were then evaluated based on the calibration and validation errors. The results show that the IDM had the overall best performance for the sampled Shanghai drivers with the lowest errors in both calibration and validation phases.

Although the number of drivers investigated in this study was relatively small, the total 60 participants were randomly sampled from the population of licensed Shanghai drivers, and the distribution of gender, age, and driving experience of the sample conformed to that of the Chinese driving population. Additionally, techniques, such as cross-validation and different measures of performance, were applied to maximize usage of the data. Therefore, the results of this study can shed some light on the driving behavior of the Chinese population in large cities.

The calibration errors of the five tested models ranged from 19% to 25%, which agrees with earlier studies of the same models (Kesting and Treiber, 2008; Brockfeld et al., 2005). The validation process generated errors from 19% to 34%. Generally, there was an increase in errors from calibration to validation. Moreover, collisions were observed in the validation phase. These results justify the necessity of using validation data to evaluate car-following models.

The best performing IDM had an error of 19% in both calibration and validation phases. As argued by Brockfeld et al. (2005), from the microscopic point of view, errors between approximately 15 and 25% probably cannot be suppressed no matter which car-following model is used. These errors are due to an exceptionally stochastic, or random, component in driver behavior (i.e., some of the behavior has no perceptible pattern and cannot be predicted). Therefore, the IDM is deemed suitable for describing Shanghai drivers' car-following behavior.

However, the Wiedemann model is used by the most popular microscopic traffic simulation tool in China (VISSIM®) to control the simulated vehicles' longitudinal movements. Compared to the Wiedemann, the IDM showed several advantages in terms of simulation. First, the IDM had significantly smaller validation errors than the Wiedemann, which means that once calibrated, it should perform better than the Wiedemann in traffic situations that did not occur in the calibration phase. Second, the IDM had a smaller standard deviation of errors than the Wiedemann, which means that the IDM has a more stable performance among different drivers. Third, the IDM has a much smaller number of parameters than the Wiedemann (5 vs. 11) and thus can be calibrated more easily and efficiently. These advantages suggest that the IDM is a more suitable longitudinal control model for microscopic traffic simulation tools in Shanghai and likely in other regions of China.

The GHR and FVD models had high validation errors, compared to their low calibration errors. This might have resulted from the large number of collisions they produced. Fig. 6 presents a case event where the lead and following vehicles came to a full stop. As can be seen, the GHR and FVD models resulted in collisions, while the IDM model could handle the highly dynamic situation well. This indicates that future improvement can be made for the GHR and FVD models concerning the better handling of dynamic situations.

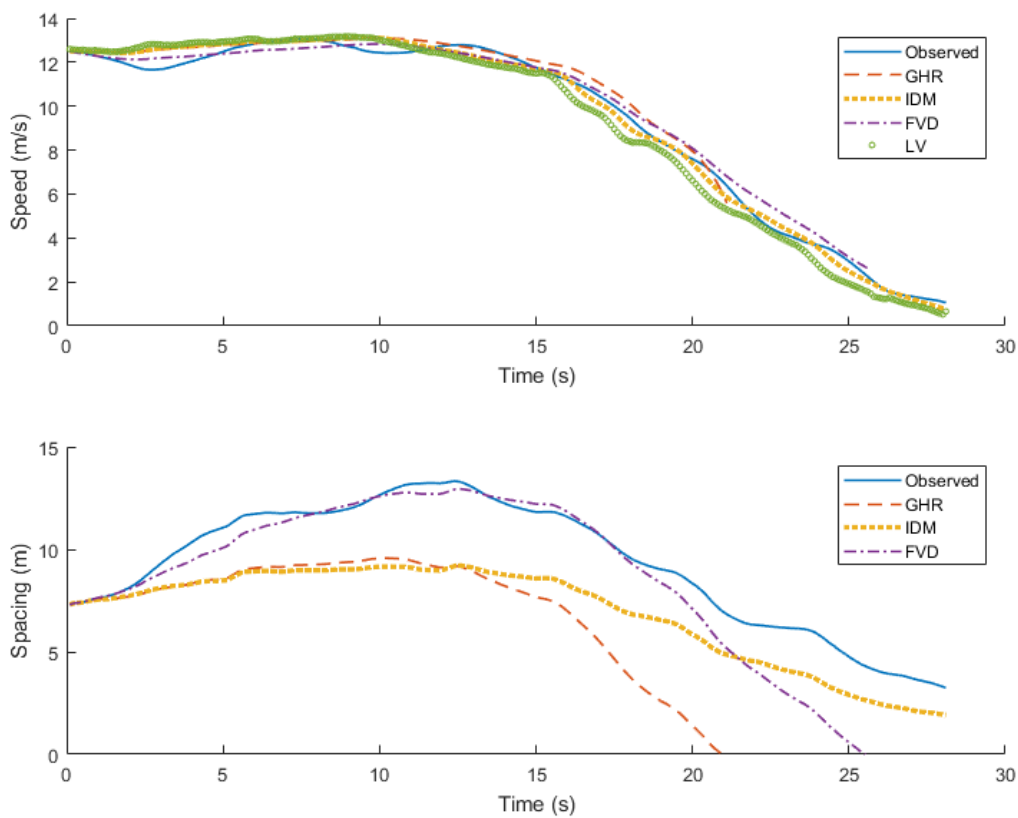

Fig. 6. A case event in which the GHR and FVD models produced collisions.

To quantitatively assess the inter-driver variability of car-following behavior, inter-driver validation was conducted with the IDM, the highest performing of the models. The IDM was first calibrated for an investigated driver and was then applied to each of the remaining 41 drivers. Therefore, two 42×42 error matrices (one for FV speed and the other for spacing) were obtained, as shown in Fig. 7. Some driver pairs had high inter-driver validation errors, while driver pairs along the diagonals generally had low errors. The average value of inter-driver spacing validation errors (31%) was much higher than intra-driver spacing validation errors (19%), which indicates that considerable behavioral differences exist between drivers, as discussed by Ossen and Hoogendoorn (2011). Moreover, five collisions occurred in the inter-driver validation phase, suggesting that inter-driver heterogeneity in car-following should not be neglected in microscopic traffic simulation, and also that simulation should be based on individual driver-specific car-following behavior. This finding identifies the need for the development of archetypes various drivers to build a traffic mix in simulation.

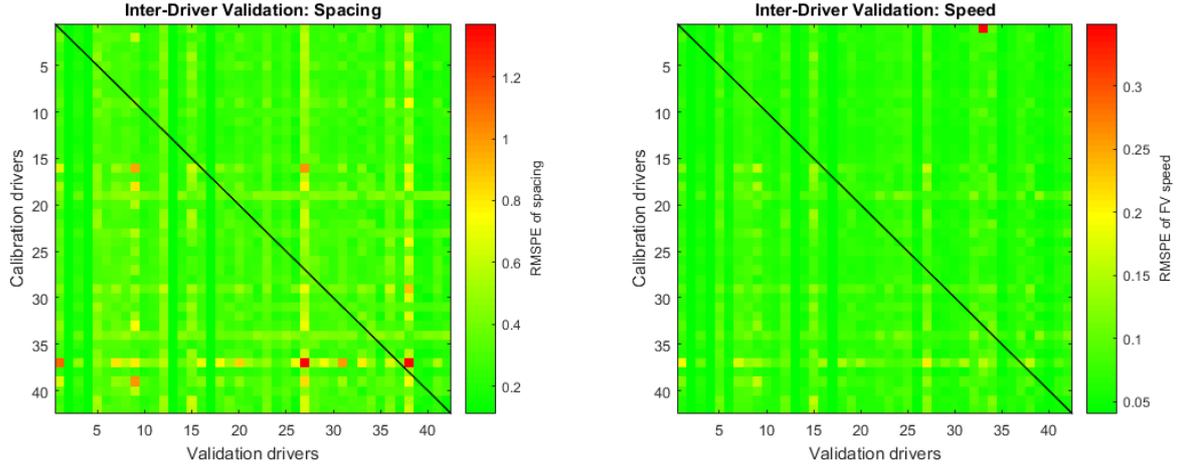

Fig. 7. IDM inter-driver validation errors for all pairs of drivers (averaged across the five iterations of cross-validation).

To further unveil the meaning of model calibration, IDM parameters estimated by calibration were compared with those by empirical observation, as shown in Fig. 8. Spacing validation errors were also compared when using calibrated and observed parameters. Further information about parameter estimation based on empirical observation is provided in Appendix D.

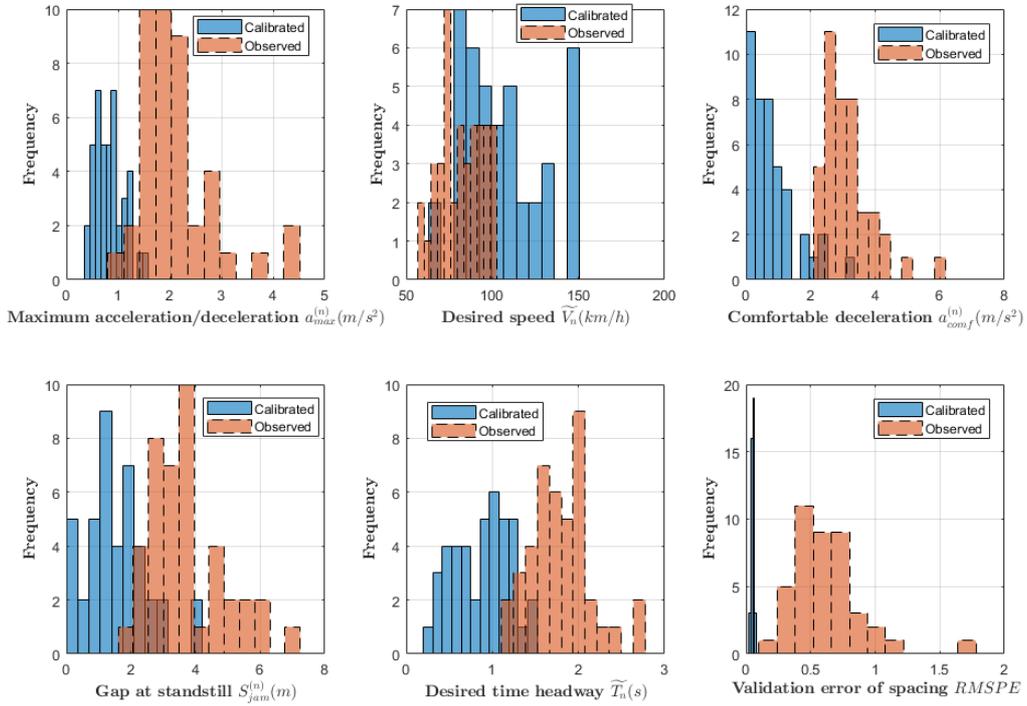

**(a) Histogram of parameters and validation errors**

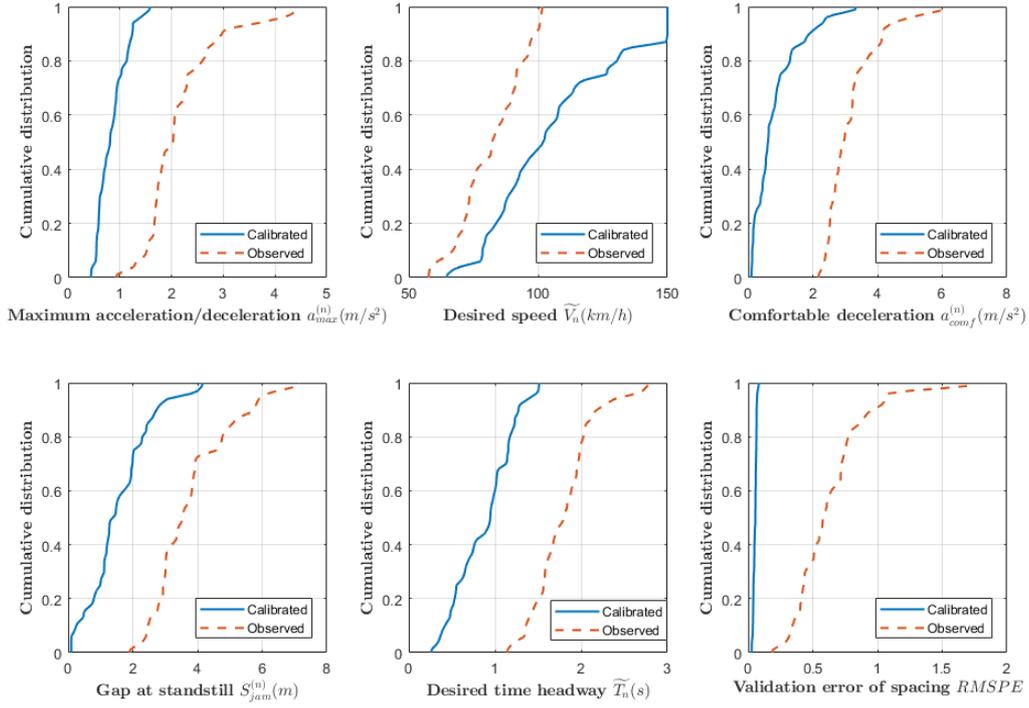

**(b) Cumulative distribution of parameters and validation errors**

Fig. 8. Comparison of calibrated and observed IDM parameters: (a) Histogram and (b) Cumulative distribution curve.

The mean and standard deviation of each IDM parameter obtained by calibration and by the empirical observations are shown in Table 4. The acceleration exponent cannot be measured directly and was assumed as 4 according to the original IDM paper (Treiber et al., 2000). A Pearson Linear Correlation analysis was conducted to test whether these two sets of parameters were correlated.

Table 4 Comparison of calibrated and observed IDM parameters.

| Parameter (unit) | Description | Calibrated | | Observed | | Pearson Correlation |
|---|---|---|---|---|---|---|
| | | Mean | Std. dev. | Mean | Std. dev. | |
| $\tilde{V}_n$ (km/h) | Desired speed | 105.62 | 24.7375 | 82.109 | 12.0748 | 0.3277 |
| $\tilde{T}_n$ (s) | Desired time headway | 0.8780 | 0.3343 | 1.7956 | 0.3488 | 0.4363 |
| $a_{max}^{(n)}$ (m/s²) | Maximum acceleration | 0.8418 | 0.2775 | 2.1384 | 1.7144 | 0.3287 |
| $a_{comf}^{(n)}$ (m/s²) | Comfortable deceleration (absolute value) | 0.8150 | 0.7251 | 3.1474 | 0.7883 | –0.0693 |
| $\beta$ | Acceleration exponent | 3.5238 | 3.6240 | 4 | 0 | — |
| $S_{jam}^{(n)}$ (m) | Standstill gap | 1.5554 | 0.9633 | 3.1474 | 1.1736 | 0.2114 |

What is noticeable from the above comparison is that 1) observed parameters generally correlated with, and distributed similarly with calibrated ones, but had higher values than calibrated ones (except for desired speed); 2) spacing validation errors were significantly larger when observed parameters were used, compared to using calibrated ones. These results may indicate that 1) interpretable calibrated model parameters are linked with corresponding observable parameters in real world, but they are not necessarily numerically equivalent (Wagner et al., 2016); and 2) parameters that can be measured in reality also need to be calibrated if better trajectory reproducing capability are to be achieved (Punzo et al., 2015).

The observed difficulty in replicating the relationships between the observable model parameters such as preferred speeds and maximum acceleration rates and the measured traffic behavior such as spacing and flow rates indicates the limitations of the investigated car-following models when applied to Shanghai drivers. This may pose a major problem for the end user of these models because the adherence of the interpretable model parameters to the reality is as important as of the simulation results. One mitigation borrowed from non-linear statistical modeling is introducing additional calibration parameters. A model with increased parametrization could be then calibrated by setting the new parameters to match the observed car-following behavior while the interpretable parameters are kept at the observed values. This approach would preserve the realism of the model through the consistency of its parameters and results.

This is the first study that utilized a massive volume of real-world Chinese driving data to calibrate, validate, and cross-compare car-following models representative of the most commonly employed types. The results of this study will be valuable for calibration, validation and development of car-following models, and for gaining a better understanding of car-following model calibration.

**Acknowledgements**

This study was jointly sponsored by the Chinese National Science Foundation (51522810), the Science and Technology Commission of Shanghai Municipality, China (18DZ1200200), the China National Engineering Laboratory for Integrated Optimization of Road Traffic and Safety Analysis Technologies, and the China 111 Project (B17032).

# Appendices

*A. Wiedemann 99 model*

The Wiedemann 99 model considers four driving regimes: free driving, closing in, following, and emergency braking. The model explicitly identifies each of these regimes according to certain thresholds, as shown in Fig. A.1 where $\Delta X$ and $\Delta V$ are, respectively, the distance and speed difference (LV minus FV) between the lead and following vehicle.

Depending on the driving regime, the driver's reaction to the stimuli coming from the lead vehicle will vary. In the current study, the acceleration rate was used as the model output. Fig. A.2 shows the calculation process of the acceleration, $a_n(t+1)$ in various driving regimes according to the inputs in the current step *t*.

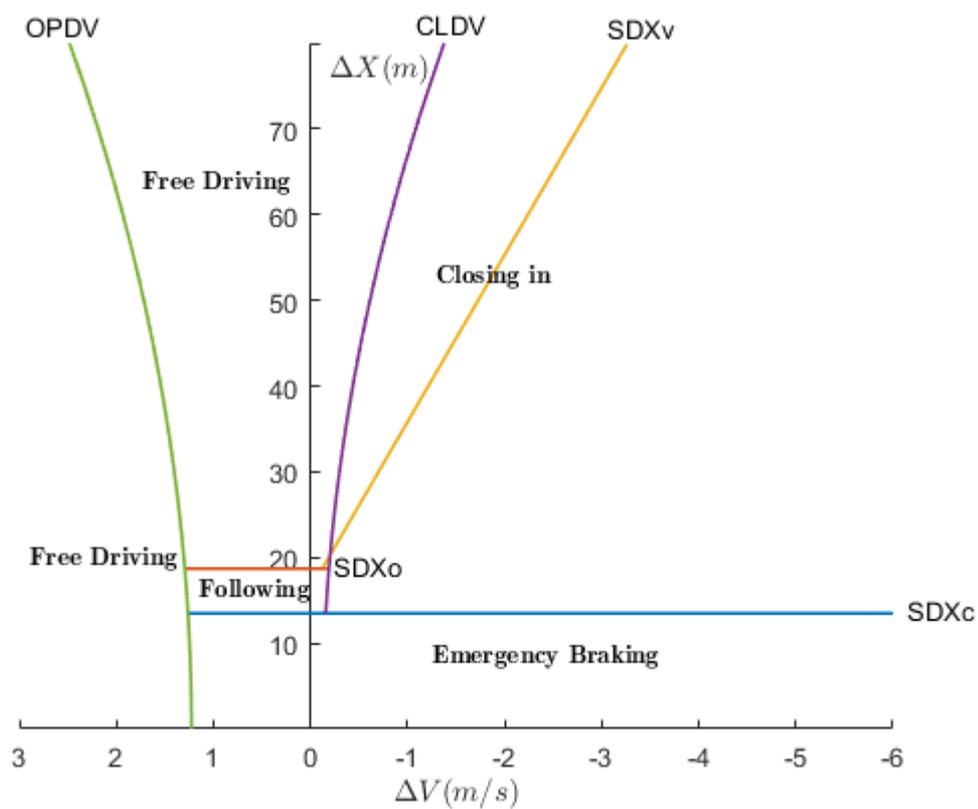

Fig. A.1. Schematic of Wiedemann model with various driving regimes and thresholds (generated based on median values of the model parameters in Table A.1).

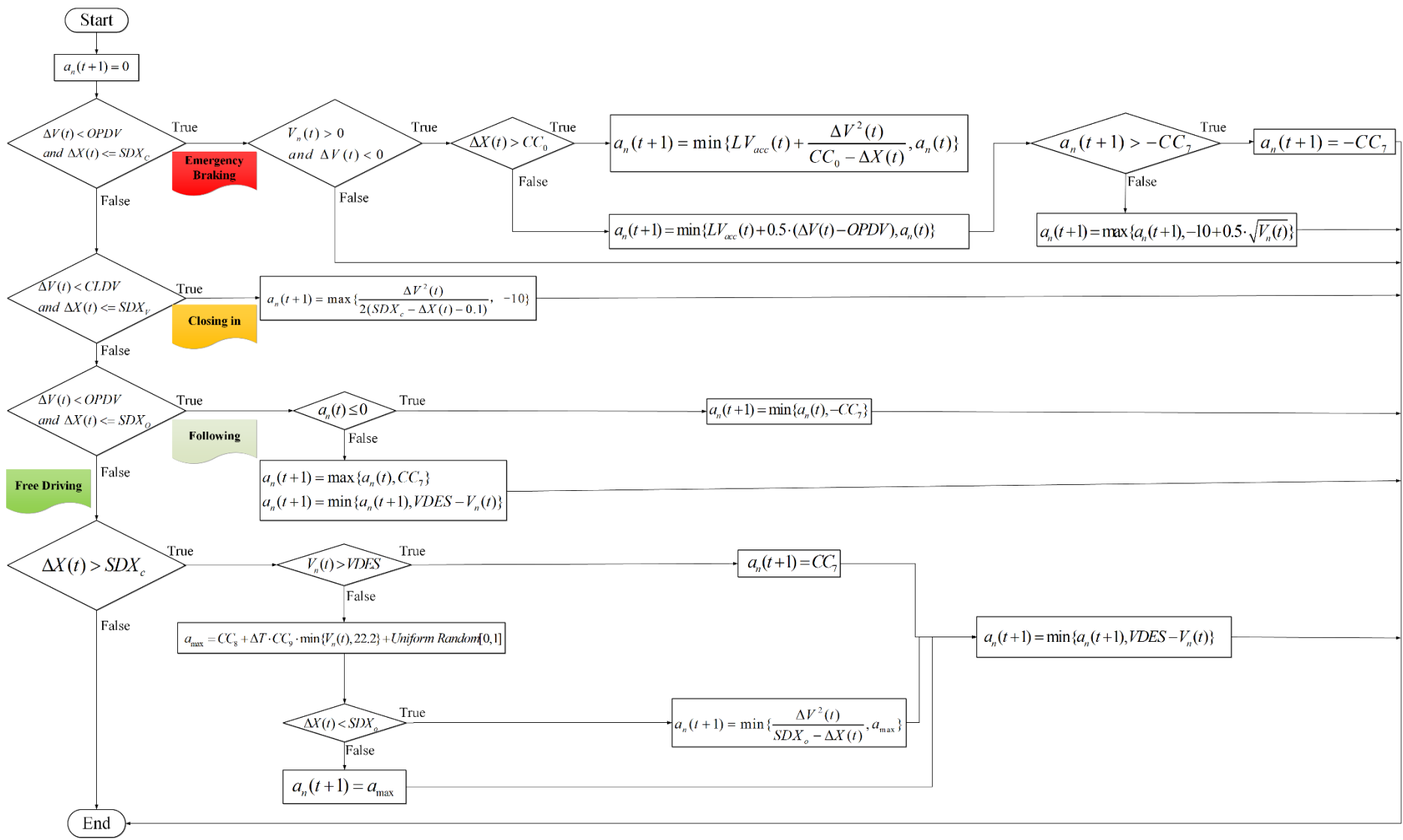

Fig. A.2. Calculation process of acceleration in the Wiedemann model (adapted from Vortish, 2014).

The corresponding thresholds in Fig. A.1 and Fig. A.2 are formulated as follows

$$\Delta X(t) = X_{n-1}(t) - X_n(t) - L_{n-1} \tag{A.1}$$

$$\Delta V(t) = V_{n-1}(t) - V_n(t) \tag{A.2}$$

$$SDX_c = CC_0 + CC_1 \cdot V_{slower} \tag{A.3}$$

$$V_{slower} = \begin{cases} V_n(t) & \text{if } \Delta V(t) > 0 \text{ or } LV_{acc}(t) < -1 m/s^2 \\ V_{n-1}(t) - \Delta V(t) \cdot RND & \text{otherwise} \end{cases} \tag{A.4}$$

$$RND = Uniform\ Random[-0.5, 0.5] \tag{A.5}$$

$$SDV = CC_6 \cdot (\Delta X(t) - L_{n-1})^2 \tag{A.6}$$

$$SDX_o = SDX_c + CC_2 \tag{A.7}$$

$$SDX_v = SDX_o + CC_3(\Delta V(t) - CC_4) \tag{A.8}$$

$$CLDV = \begin{cases} -SDV + CC_4 & \text{if } V_{n-1}(t) > 0 \\ 0 & \text{otherwise} \end{cases} \tag{A.9}$$

$$OPDV = \begin{cases} SDV + CC_5 & \text{if } V_n(t) > CC_5 \\ SDV & \text{otherwise} \end{cases} \tag{A.10}$$

where

$X_n, X_{n-1}$ = position of the following and lead vehicle respectively,

$\Delta X$ = following gap (m),

$V_n, V_{n-1}$ = velocity of the following and lead vehicle respectively (m/s),

$L_{n-1}$ = length of the lead vehicle (m),

$\Delta V$ = speed difference between the following and lead vehicle (i.e., speed of following vehicle minus speed of lead vehicle), (m/s),

$SDX_c$ = minimum safe following distance (m),

$LV_{acc}$ = acceleration of the lead vehicle (m/s²),

$SDV$ = perception threshold of speed difference (m/s),

$SDX_o$ = maximum following distance (m),

$SDX_v$ = distance threshold of following vehicle perceiving its approach to a slower leader (m),

$CLDV$ = perception threshold of speed difference at short decreasing distances (m/s),

$OPDV$ = perception threshold of speed difference at short but increasing distances (m/s),

$VDES$ = desired speed of following vehicle (km/h),

$a_{max}$ = maximum acceleration of vehicle (m/s²), and

$CC_0$ to $CC_9$ = model parameters.

The constraints applied to the Wiedemann 99 parameters, according to Vortish, 2014, are shown in Table A.1.

Table A.1 Summary of the Wiedemann parameters.

| Parameter (unit) | Short description | Bounds | Mean | Median | Std. dev. | 5% | 95% |
|---|---|---|---|---|---|---|---|
| $CC_0$ (m) | Standstill gap | [0, 20] | 1.0192 | 0.6306 | 0.9696 | 0.1056 | 2.5673 |
| $CC_1$ (m) | Headway time | [0, 5] | 1.4242 | 1.3750 | 0.4326 | 0.8200 | 2.1980 |
| $CC_2$ (m) | 'Following' variation | [0, 10] | 5.8715 | 5.2325 | 3.9500 | 0.5208 | 9.9986 |
| $CC_3$ (s) | Threshold for entering 'following' | [–20, 0] | -17.1579 | -19.4882 | 4.7313 | -19.970 | -4.5030 |
| $CC_4$ (m/s) | Negative 'following' threshold | [–5, 0] | -0.2312 | -0.1215 | 0.2604 | -0.9616 | -0.1006 |
| $CC_5$ (m/s) | Positive 'following' threshold | [0.1, 5] | 1.7946 | 1.2263 | 1.5706 | 0.1391 | 4.7846 |
| $CC_6$ ($10^{-4}$ rad/s) | Speed dependency of oscillation | [0.1, 20] | 3.5519 | 1.9640 | 3.8974 | 0.2664 | 11.0473 |
| $CC_7$ (m/s$^2$) | Oscillation acceleration | [–1, 1] | 0.5350 | 0.5379 | 0.3343 | 0.0974 | 0.9729 |
| $CC_8$ (m/s$^2$) | Standstill acceleration | [0, 8] | 4.0101 | 3.9712 | 3.1826 | 0.1087 | 7.9984 |
| $CC_9$ (m/s$^2$) | Acceleration at 80 km/h | [0, 8] | 2.6549 | 0.6971 | 3.0612 | 0.1095 | 7.8852 |
| VDES | Desired speed of FV | [1, 150] | 86.0492 | 81.1745 | 17.8998 | 67.2759 | 124.1050 |

*B. Determining the thresholds for car-following extraction*

The criteria and thresholds for car-following extraction were defined in three steps.

**(1) Determine initial criteria to extract potential car-following periods**

Table B.1 shows a summary of the car-following extraction criteria used in previous studies. Based on these criteria, initial criteria and thresholds for potential car-following extraction were defined:

- the subject vehicle and lead vehicle are driving in the same lane, with a range less than 150 m;
- the following duration is longer than 15 s.

These initial criteria and thresholds are relatively loose, to avoid missing potential car-following periods.

Table B.1 Summary of car-following extraction criteria in previous studies

| Study | Lateral distance (m) | Range (m) | Duration (s) |
|---|---|---|---|
| LeBlanc et al. (2013) | | | >15 |
| Chong et al. (2013) | <1.9 | <120 | >30 |
| Fernandez (2011) | | <100 | |
| Current study initial | At the same lane (manual check) | <150 | >15 |

| | Current study final | <2.5 | <120 | >15 |
|---|---|---|---|---|

## (2) Sample valid car-following periods

From the potential car-following periods extracted in Step One, 155 valid car-following periods were sampled according to the definition of car-following: a situation in which a vehicle's speed and longitudinal position are influenced by the vehicle immediately ahead of it (Ranney, 1999). Specifically, speed and position curves along with video materials were applied to determine whether the subject vehicle's driving was influenced by the lead vehicle, as demonstrated in Fig. B.1.

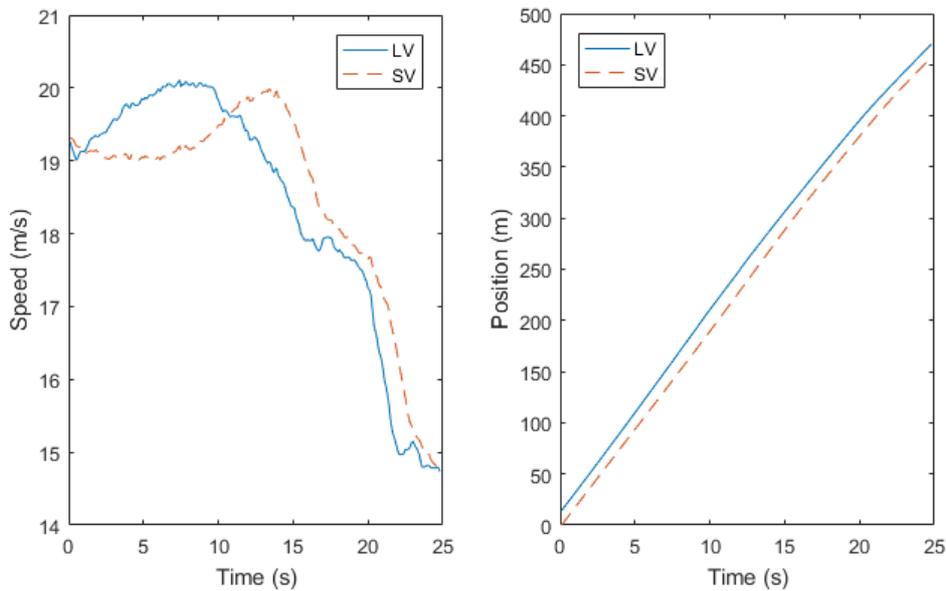

Fig. B.1. Typical speed and position curves for car-following periods.

## (3) Determine final criteria based on the sampled valid car-following periods

The parameter ranges of the sampled car-following periods are summarized in Table B.2. Based on these parameter ranges and previous studies, final criteria and thresholds for car-following extraction were determined, as shown in the last row of Table B.1.

Table B.2 Parameter ranges of the sampled car-following periods

| ID | Duration (s) | Range (m) | | Lateral distance (m) | |
|---|---|---|---|---|---|
| | | Min | Max | Min | Max |
| 1 | 27.40 | 27.52 | 34.37 | -2.37 | 2.30 |
| 2 | 15.32 | 11.14 | 17.06 | -0.51 | 0.58 |
| 3 | 36.40 | 8.58 | 24.99 | -0.64 | 1.60 |
| 4 | 35.00 | 9.66 | 19.49 | -0.58 | 2.02 |
| 5 | 37.10 | 45.02 | 52.16 | -1.09 | 1.50 |
| 6 | 25.70 | 64.03 | 81.01 | -2.02 | 1.70 |
| 7 | 22.80 | 10.27 | 30.37 | -1.82 | 0.74 |
| 8 | 25.20 | 9.47 | 34.05 | -1.25 | 0.90 |
| 9 | 46.30 | 17.57 | 58.34 | -1.28 | 1.73 |
| 151 | 24.00 | 10.75 | 28.35 | -0.93 | 1.09 |
| 152 | 25.20 | 7.55 | 14.75 | -0.80 | 0.96 |
| 153 | 116.40 | 8.48 | 24.96 | -0.48 | 2.24 |
| 154 | 23.40 | 15.78 | 25.12 | -1.34 | 0.83 |
| 155 | 34.80 | 34.46 | 45.92 | -0.93 | 2.05 |
| Aggregate | **15.20** | 7.52 | **113.00** | **-2.37** | 2.30 |

*C. Cumulative distribution of parameter estimates for the models*

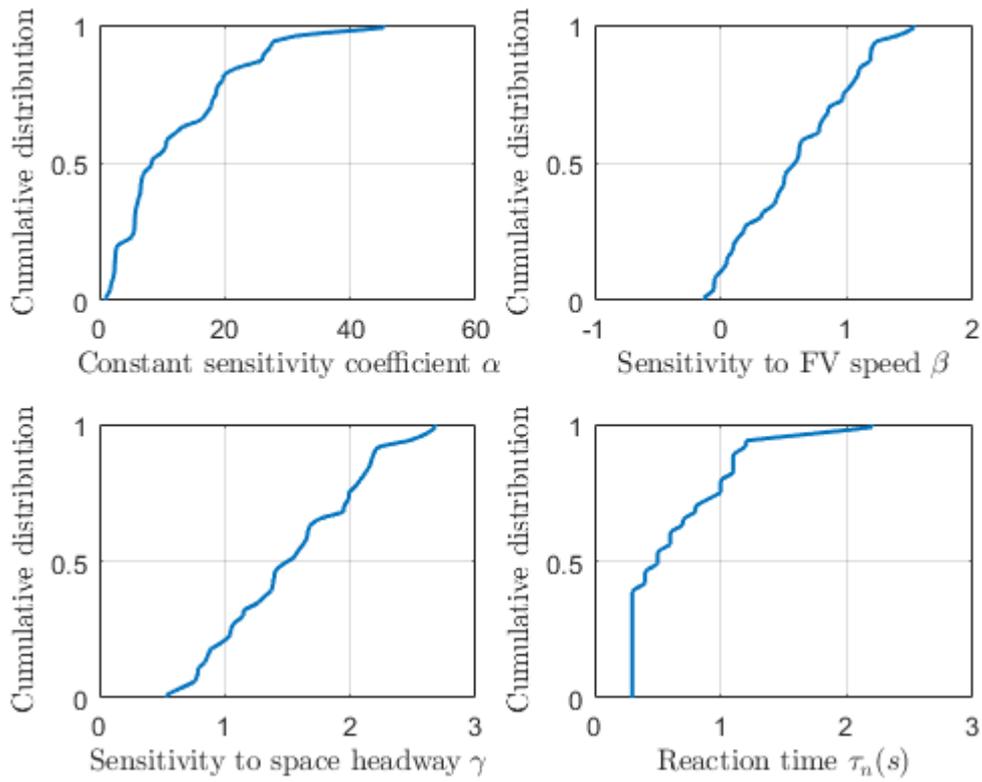

Fig. C.1. Cumulative distribution of parameter estimates for the GHR model.

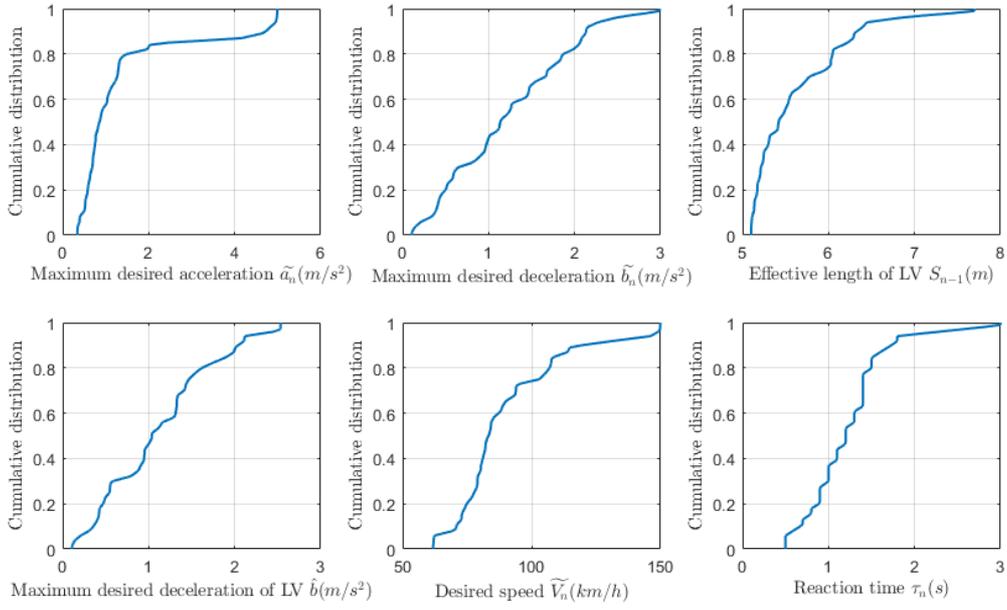

Fig. C.2. Cumulative distribution of parameter estimates for the Gipps model.

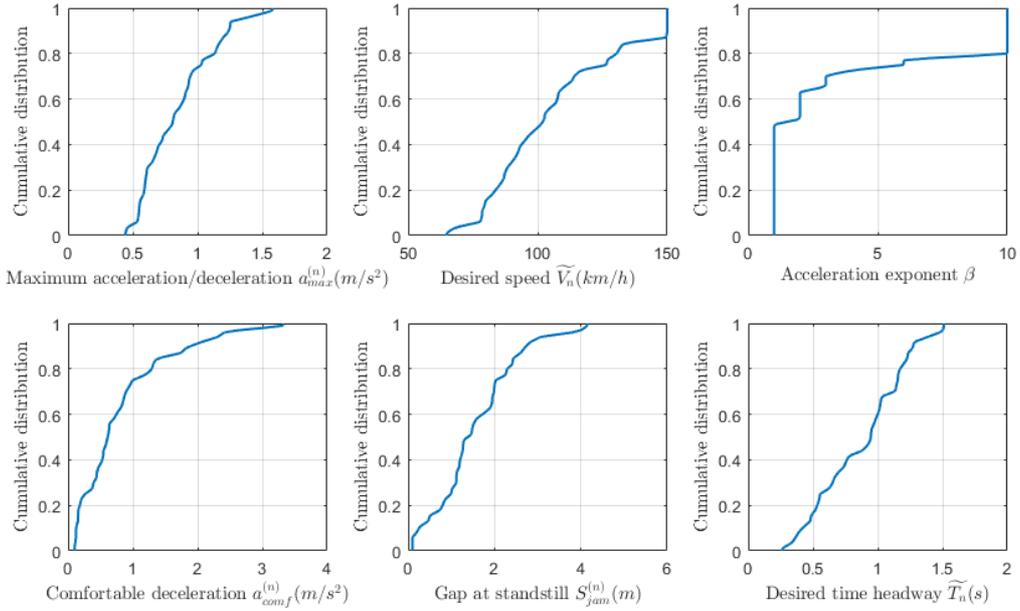

Fig. C.3. Cumulative distribution of parameter estimates for the IDM.

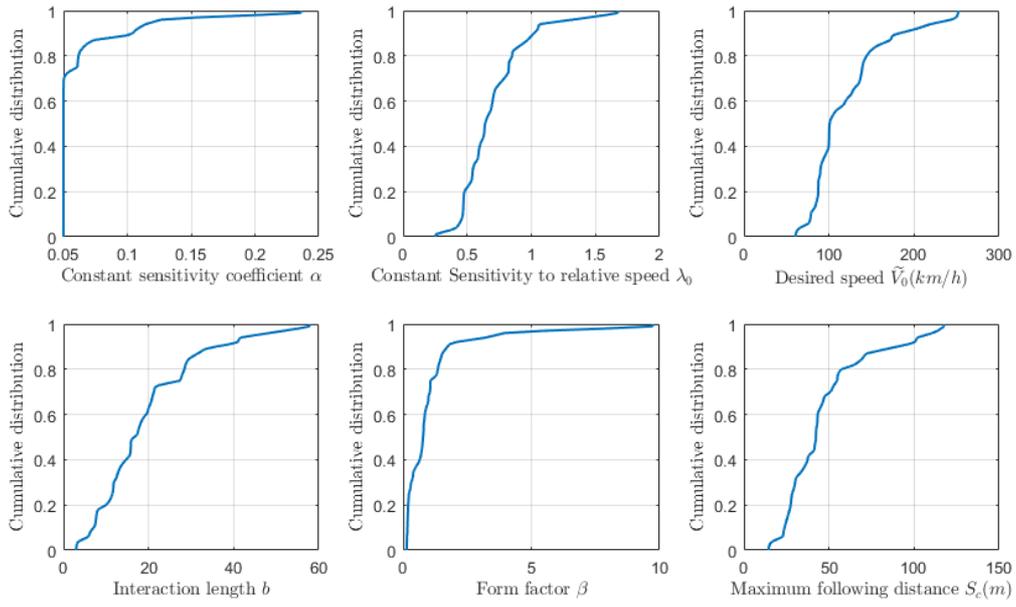

Fig. C.4. Cumulative distribution of parameter estimates for the FVD model.

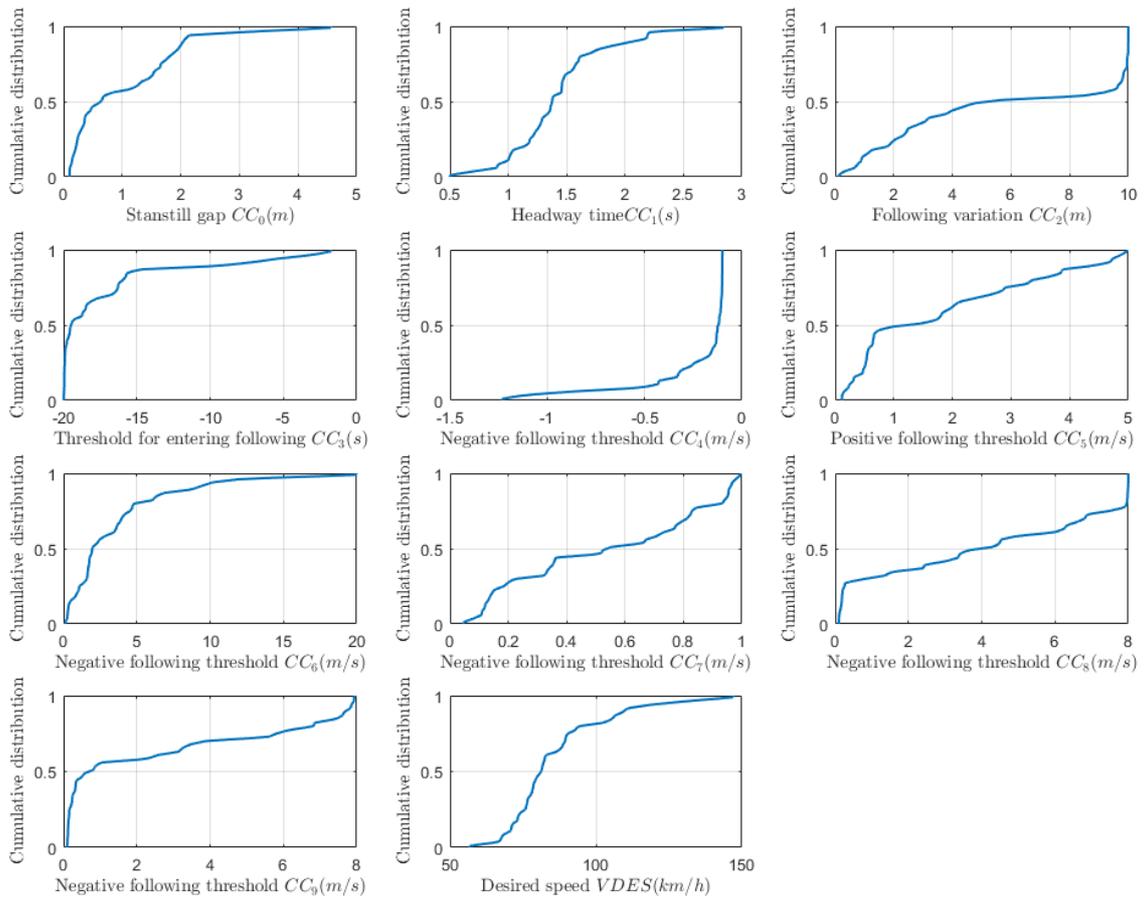

Fig. C.5. Cumulative distribution of parameter estimates for the Wiedemann 99 model.

*D. Estimating IDM parameter by empirical observation*

The meaning of IDM's six parameters is intuitive and each parameter relates to a certain driving regime (Treiber and Kesting, 2013): the desired speed is relevant for cruising in free-traffic conditions, the desired time headway pertains to steady-state car-following, the standstill gap to creeping and standing traffic, and the acceleration exponent, maximum acceleration and comfort acceleration relate to non-steady-state traffic flow. Therefore, theses parameters were estimated for each driver by investigating the empirical data of the corresponding driving regimes (the acceleration exponent cannot be measured directly and was assumed as 4 according to the original IDM paper):

- Desired speed: free-driving data with spacing >120 m (according to Appendix B) were selected from a driver's all urban expressway driving events, and the averaged following vehicle speed was used as the desired speed.
- Desired time headway: steady-state car-following data with absolute values of relative speed < 1 m/s were selected from a driver's urban expressway car-following events, and the mean time headway was used as the desired time headway (Dey and Chandra, 2009).
- Standstill gap: standing-traffic data with following vehicle speed < 1 m/s were selected from a driver's car-following events, and the averaged following gap was used as the standstill gap.
- Maximum acceleration: the maximum acceleration rate a driver adopted during his/her all car-following events (Sangster et al., 2013).
- Comfortable deceleration: the maximum deceleration rate a driver adopted during his/her all car-following events (Sangster et al., 2013).